\documentclass[sigconf]{aamas}  
\usepackage{booktabs}

\setcopyright{ifaamas}  
\acmDOI{}  
\acmISBN{}  
\acmConference[AAMAS'18]{Proc.\@ of the 17th International Conference on Autonomous Agents and Multiagent Systems (AAMAS 2018)}{July 10--15, 2018}{Stockholm, Sweden}{M.~Dastani, G.~Sukthankar, E.~Andr\'{e}, S.~Koenig (eds.)}  
\acmYear{2018}  
\copyrightyear{2018}  
\acmPrice{}  

\usepackage{amssymb}
\usepackage{amsmath}

\makeatletter
\g@addto@macro\normalsize{%
  \setlength\abovedisplayskip{0pt}
  \setlength\belowdisplayskip{0pt}
  \setlength\abovedisplayshortskip{0pt}
  \setlength\belowdisplayshortskip{0pt}
}
\makeatother

\usepackage{bm}

\usepackage{amsthm}
\usepackage{xcolor}
\usepackage{bbm}
\usepackage{graphicx}
\usepackage[font=scriptsize]{caption}
\usepackage{subfigure}
\usepackage{url}
\usepackage{threeparttable}
\usepackage{multirow}
\usepackage[linesnumbered,vlined,ruled]{algorithm2e}
\SetKwProg{Fn}{Function}{}{}

\usepackage{tabularx, booktabs}

\theoremstyle{plain}

\newtheorem{thm}{Theorem}

\makeatletter
\def\thm@space@setup{%
  \thm@preskip=\parskip \thm@postskip=0pt
}
\makeatother

\begin{document}

\title{Multi-Agent Path Finding with Deadlines: Preliminary Results}  
\titlenote{Hang Ma, Jiaoyang Li, T. K. Satish Kumar, and Sven Koenig are affiliated with the University of Southern California (USC). The research at USC was supported by the National Science Foundation (NSF) under grant numbers 1724392, 1409987, and 1319966 as well as a gift from Amazon.}

\subtitle{Extended Abstract}





\author{Hang Ma}
\affiliation{%
  \institution{USC}
}
\authornotemark[1]
\email{hangma@usc.edu}

\author{Glenn Wagner}
\affiliation{%
  \institution{CSIRO}
}
\author{Ariel Felner}
\affiliation{%
  \institution{Ben-Gurion University}
}
\author{Jiaoyang Li}\authornotemark[1]
\author{T. K. Satish Kumar}\authornotemark[1]
\author{Sven Koenig}\authornotemark[1]

\providecommand{\agent}[1]{a_{#1}} 
\providecommand{\config}{q}
\providecommand{\tstep}{t}
\providecommand{\tcrit}{T_{end}}
\providecommand{\dtnode}[1]{\ensuremath{N_{#1}}}
\providecommand{\nodedead}[1]{\dtnode{#1}.\emph{dead}}
\providecommand{\nodelive}[1]{\dtnode{#1}.\emph{live}}
\providecommand{\nodecost}[1]{\dtnode{#1}.\emph{cost}}
\providecommand{\group}{\gamma}

\begin{abstract}  
We formalize the problem of multi-agent path finding with deadlines (MAPF-DL).
The objective is to maximize the number of agents that can reach their given
goal vertices from their given start vertices within a given deadline, without
colliding with each other. We first show that the MAPF-DL problem is NP-hard
to solve optimally. We then present an optimal MAPF-DL algorithm based on a
reduction of the MAPF-DL problem to a flow problem and a subsequent compact
integer linear programming formulation of the resulting reduced abstracted
multi-commodity flow network.
\end{abstract}

\maketitle

\section{Introduction}
Multi-agent path finding (MAPF) is the problem of planning collision-free
paths for multiple agents in known environments from their given start
vertices to their given goal vertices. MAPF is important, for example, for
aircraft-towing vehicles \cite{airporttug16}, warehouse and office robots
\cite{kiva,DBLP:conf/ijcai/VelosoBCR15}, and game characters \cite{MaAIIDE17}.
The objective is to minimize the sum of the arrival times of the agents or the
makespan. The MAPF problem is NP-hard to solve optimally \cite{YuLav13AAAI}
and even to approximate within a small constant factor for makespan
minimization \cite{MaAAAI16}. It can be solved with reductions to other
well-studied combinatorial problems
\cite{Surynek15,YuLav13ICRA,erdem2013general,GTAPF} and dedicated optimal
\cite{ODA,ODA11,EPEJAIR,DBLP:journals/ai/SharonSGF13,MStar,DBLP:journals/ai/SharonSFS15,ICBS,FelnerICAPS18},
bounded-suboptimal \cite{ECBS,CohenUK16}, and suboptimal MAPF algorithms
\cite{WHCA,WHCA06,WangB11,PushAndSwap,PushAndRotate,DBLP:conf/iros/WagnerC11},
as described in several surveys \cite{MaWOMPF16,SoCS2017Surv}.

The MAPF problem has recently been generalized in different directions
\cite{MaAAMAS16,HoenigICAPS16,MaWOMPF16,HoenigIROS16,MaAAAI17,MaAAMAS17,MaIEEE17} but
none of them capture an important characteristic of many applications, namely
the ability to meet deadlines. We thus formalize the multi-agent path finding
problem with deadlines (MAPF-DL problem). The objective is to maximize the
number of agents that can reach their given goal vertices from their given
start vertices within a given deadline, without colliding with each other. In
previously studied MAPF problems, all agents have to be routed from their
start vertices to their goal vertices, and the objective is with regard to
resources such as fuel (sum of arrival times) or time (makespan). In the
MAPF-DL problem, on the other hand, the resources are the agents
themselves. We first show that the MAPF-DL problem is NP-hard to solve
optimally. We then present an optimal MAPF-DL algorithm based on a reduction
of the MAPF-DL problem to a flow problem and a subsequent compact
integer linear programming formulation of the resulting reduced abstracted
multi-commodity flow network.

\section{MAPF-DL Problem}

We formalize the MAPF-DL problem as follows: We are given a \emph{deadline},
denoted by a time step $\tcrit$, an undirected graph $G = (V,E)$, and $M$
agents $\agent{1}, \agent{2} \ldots \agent{M}$. Each agent $\agent{i}$ has a
start vertex $s_i$ and a goal vertex $g_i$. In each time step, each agent
either moves to an adjacent vertex or stays at the same vertex. Let $l_i(t)$
be the vertex occupied by agent $\agent{i}$ at time step $t \in
\{0\ldots\tcrit\}$. Call an agent $a_i$ \emph{successful} iff it occupies its
goal vertex at the deadline $\tcrit$, that is, $l_i(\tcrit) = g_i$. A
\emph{plan} consists of a path $l_i$ assigned to each successful agent
$\agent{i}$. Unsuccessful agents are removed at time step zero and thus have
no paths assigned to them.\footnote{Depending on the application, the
  unsuccessful agents can be removed at time step zero, wait at their start
  vertices, or move out of the way of the successful agents. We choose the
  first option in this paper. If the unsuccessful agents are not removed, they
  can obstruct other agents. However, our proof of NP-hardness does not depend
  on this assumption, and our MAPF-DL algorithm can be adapted to other
  assumptions.}  A solution is a plan that satisfies the following conditions:
(1) For all successful agents $\agent{i}$, $l_i(0) = s_i$ [each successful
  agent starts at its start vertex]. (2) For all successful agents $\agent{i}$
and all time steps $t > 0$, $(l_i(t - 1), l_i(t))\in E$ or $l_i(t - 1) =
l_i(t)$ [each successful agent always either moves to an adjacent vertex or
  does not move]. (3) For all pairs of different successful agents $\agent{i}$
and $\agent{j}$ and all time steps $t$, $l_i(t) \neq l_j(t)$ [two successful
  agents never occupy the same vertex simultaneously]. (4) For all pairs of
different successful agents $\agent{i}$ and $\agent{j}$ and all time steps $t
> 0$, $l_i(t - 1) \neq l_j(t)$ or $l_j(t - 1) \neq l_i(t)$ [two successful
  agents never traverse the same edge simultaneously in opposite directions].
Define a \emph{collision} between two different successful agents $\agent{i}$
and $\agent{j}$ to be either a \emph{vertex collision} ($\agent{i}$,
$\agent{j}$, $v$, $t$) iff $v = l_i(t) = l_j(t)$ (corresponding to Condition
3) or an \emph{edge collision} ($\agent{i}$, $\agent{j}$, $u$, $v$, $t$) iff
$u = l_i(t) = l_j(t+1)$ and $v = l_j(t) = l_i(t+1)$ (corresponding to
Condition 4). The objective is to maximize the number of successful agents
$M_{succ} = \left\vert\{ \agent{i} | l_i(\tcrit) = g_i\}\right\vert$.

\begin{figure}[]
  \centering
  \begin{minipage}[c]{0.5\columnwidth}{
  \centering\includegraphics[width=0.7\columnwidth]{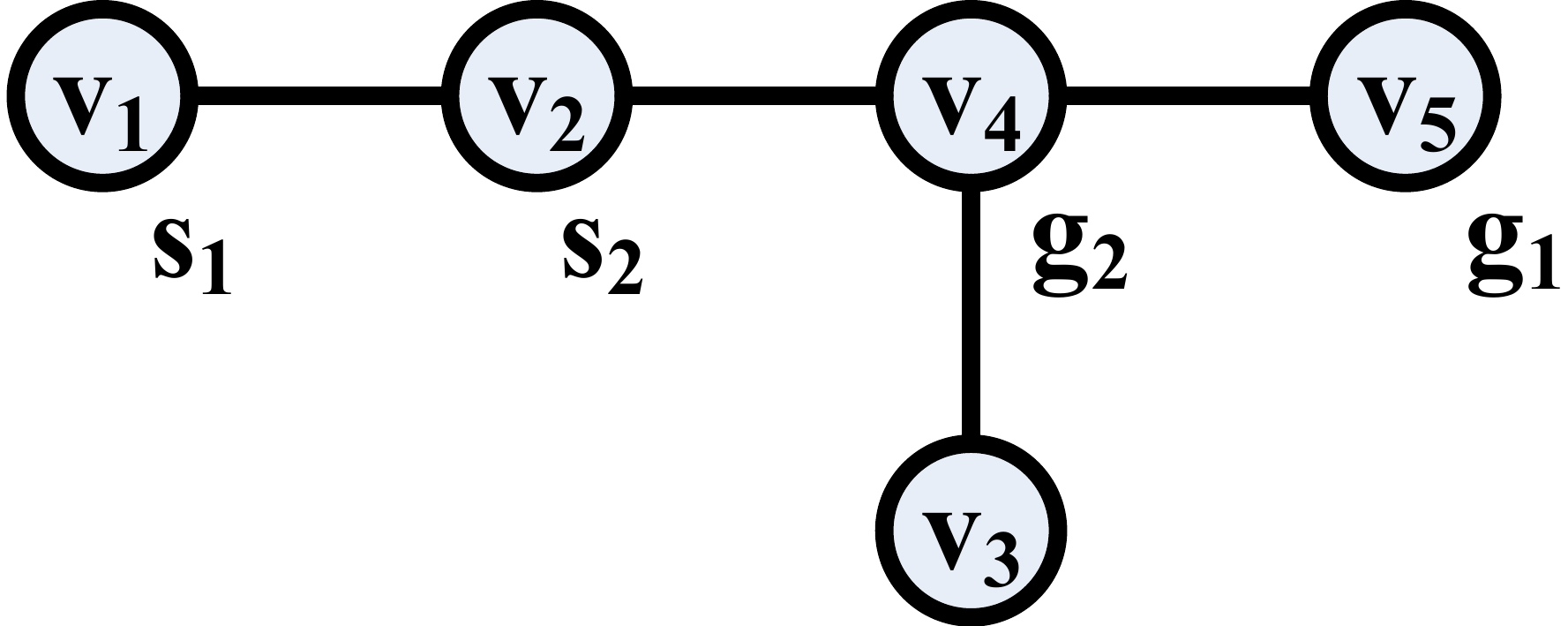}\\ \centering\scriptsize\textbf{(a)}\\
  \centering\includegraphics[width=0.7\columnwidth]{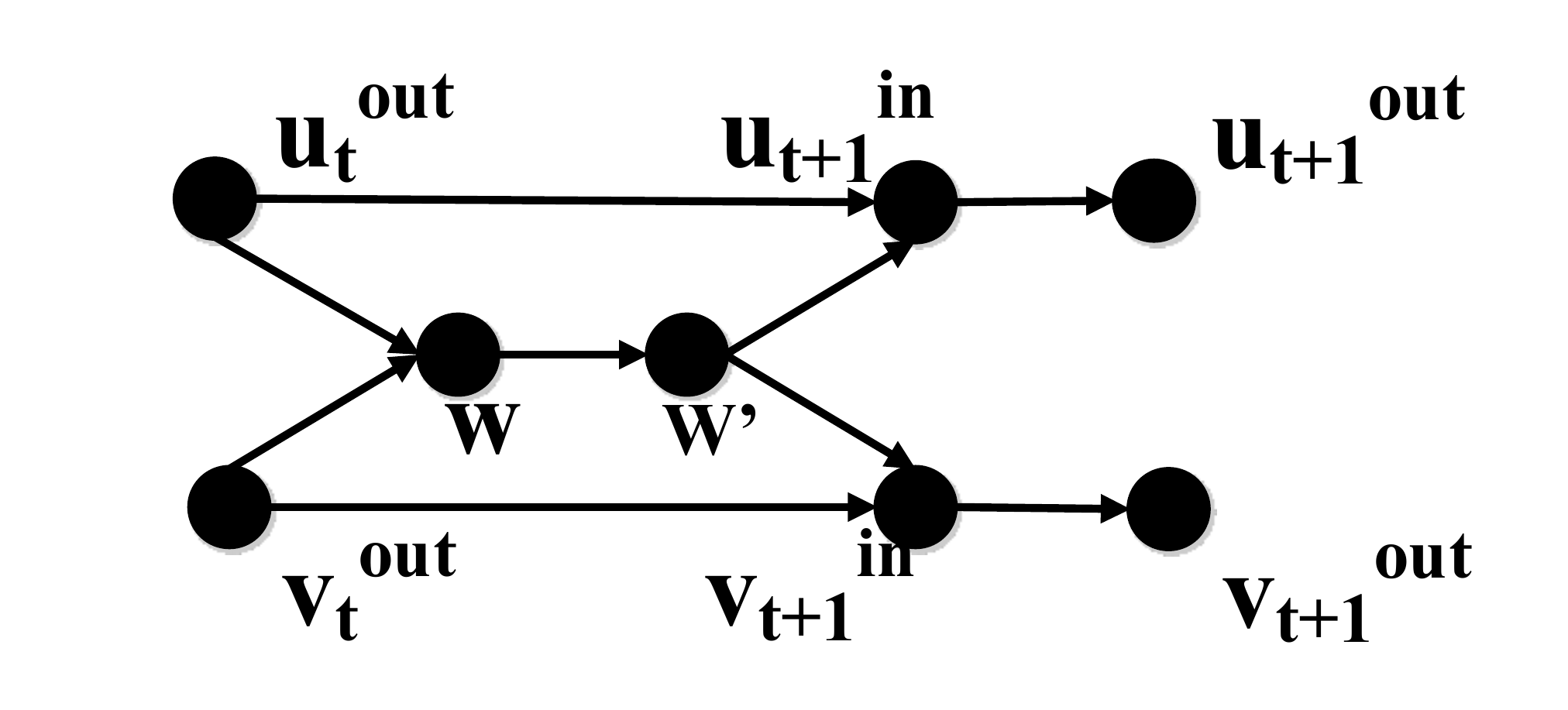}\\
  \centering\scriptsize\textbf{(b)}\\}\end{minipage}\begin{minipage}[c]{0.5\columnwidth}{
  \centering\includegraphics[width=0.6\columnwidth]{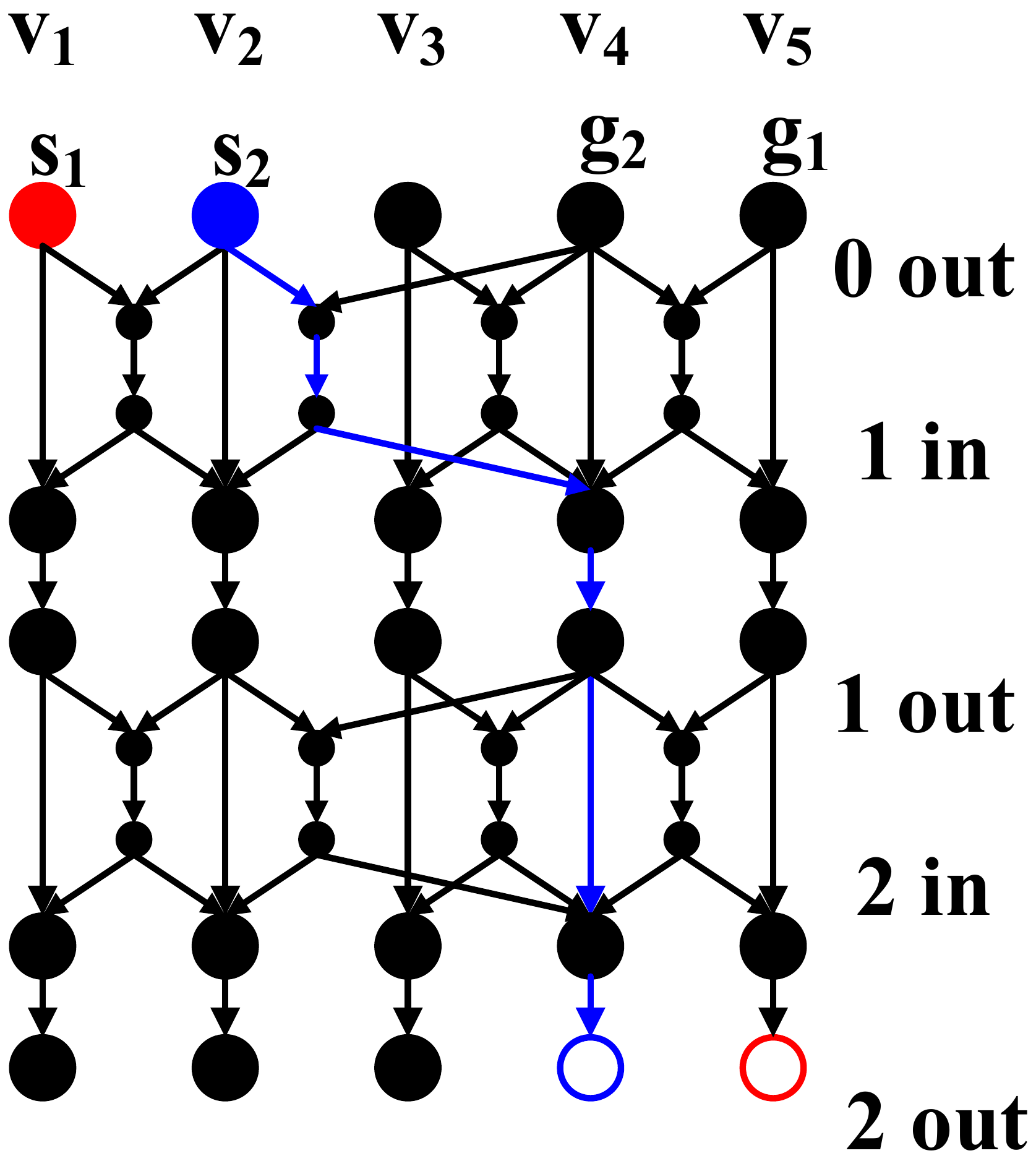}\\
  \scriptsize\textbf{(c)}\\}\end{minipage}
  \caption{(a) Running example of a MAPF-DL instance. (b) Construction for
    edge $(u,v)\in E$ between time steps $t$ and $t+1$. (c) Flow network
    for the running example.}\label{fig:example}
\end{figure}

\begin{thm}
It is NP-hard to compute a MAPF-DL solution with the maximum number of successful agents.
\end{thm}

The proof of the theorem reduces the \textbf{$\bm\le$3,$\bm=$3-SAT} problem
\cite{cat1984}, an NP-complete version of the Boolean satisfiability problem,
to the MAPF-DL problem. The reduction is similar to the one used for proving
the NP-hardness of approximating the optimal makespan for the MAPF problem
\cite{MaAAAI16}. It constructs a MAPF-DL instance with deadline $\tcrit = 3$
that has a solution where all agents are successful iff the given
\textbf{$\bm\le$3,$\bm=$3-SAT} instance is satisfiable.

\section{Optimal MAPF-DL Algorithm}

Our optimal MAPF-DL algorithm first reduces the MAPF-DL problem to the maximum
(integer) multi-commodity flow problem, which is similar to the reductions of
the MAPF and TAPF problems to multi-commodity flow problems
\cite{YuLav13ICRA,MaAAMAS16}: Given a MAPF-DL instance with deadline $\tcrit$,
we construct a multi-commodity flow network $\mathcal{N} = (\mathcal{V},
\mathcal{E})$ with vertices $\mathcal{V} = \bigcup_{v \in V} (\{v_0^{out}\}
\cup \bigcup_{t=1}^{\tcrit}\{v_t^{in},v_t^{out}\})$ and directed edges
$\mathcal{E}$ with unit capacity. The vertices $v_t^{out}$ represent vertex $v
\in V$ at the end of time step $t$ and the beginning of time step $t+1$, while
the vertices $v_t^{in}$ are intermediate vertices. For each agent $\agent{i}$,
we set a supply of one at (start) vertex $(s_i)_0^{out}$ and a demand of one
at (goal) vertex $(g_i)^{out}_{\tcrit}$, both for commodity type $i$
(corresponding to agent $a_i$). For each time step, we construct the gadgets
shown in Figure~\ref{fig:example}(b) to prevent vertex and edge
collisions. The objective is to maximize the total amount of integral flow
received in all vertices $(g_i)^{out}_{\tcrit}$, which can be achieved via a
standard integer linear programming (ILP) formulation.

\begin{thm}
\label{makespan and multiflow}
There is a one-to-one correspondence between all solutions of a MAPF-DL
instance with the maximum number of successful agents and all maximum integral
flows on the corresponding flow network.
\end{thm}

The proof of the theorem is similar to the one for the reduction of the MAPF
problem to the multi-commodity flow problem \cite{YuLav13ICRA}.

Figure~\ref{fig:example}(a) shows a MAPF-DL instance with deadline $\tcrit =
2$. Agents $\agent{1}$ and $\agent{2}$ have start vertices $s_1$ and $s_2$ and
goal vertices $g_1$ and $g_2$, respectively. The number of successful agents
is at most $M_{succ} = 1$ because only agent $\agent{2}$ can reach its goal
vertex in two time steps. Figure~\ref{fig:example}(c) shows the corresponding
flow network with a maximum flow (in color) that corresponds to a solution
with unsuccessful agent $a_1$ and successful agent $a_2$ with path $\langle
v_2$, $v_4$, $v_4 \rangle$.

\begin{figure}
  \centering
  \includegraphics[width=0.27\columnwidth]{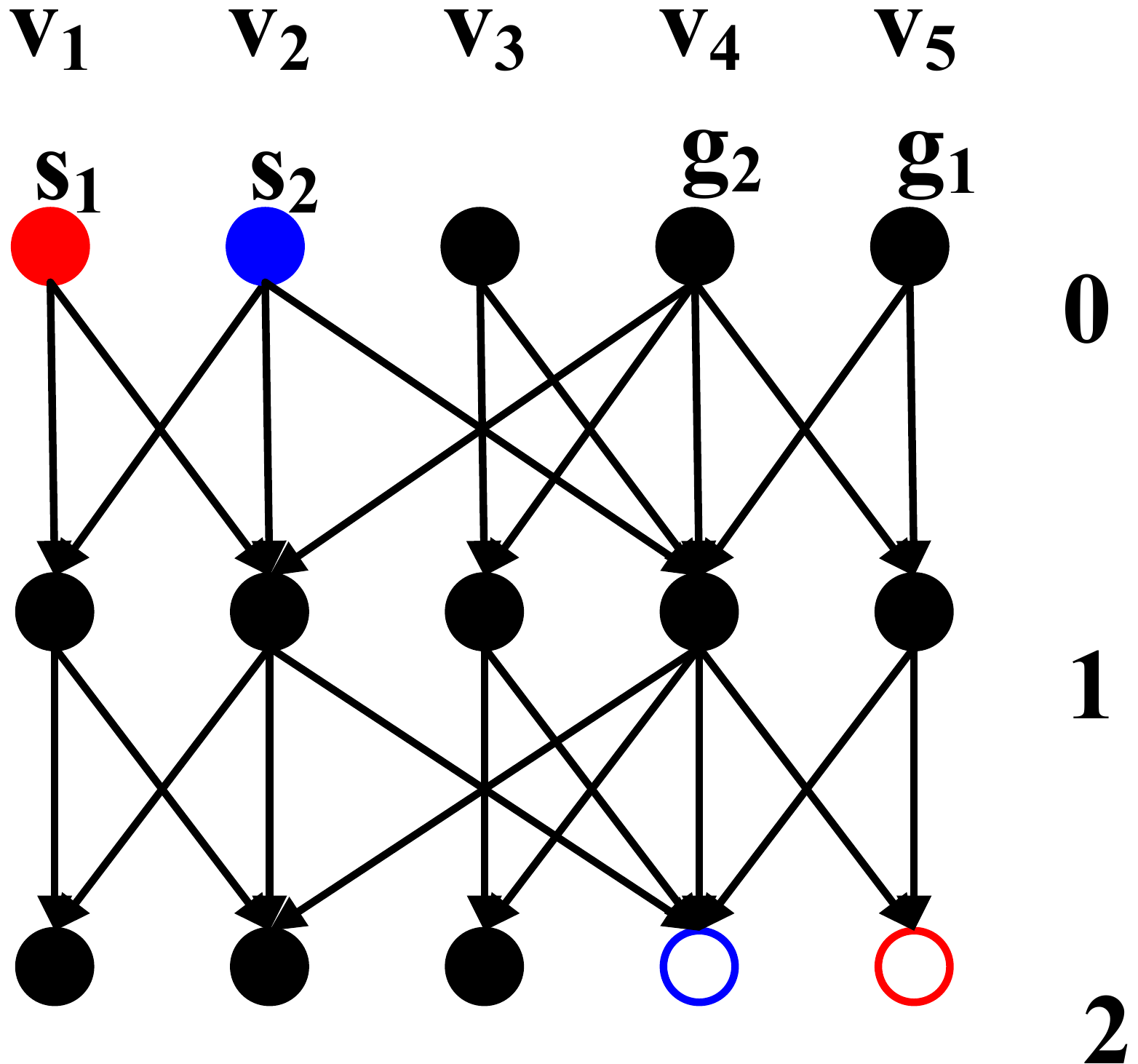}\qquad\includegraphics[width=0.27\columnwidth]{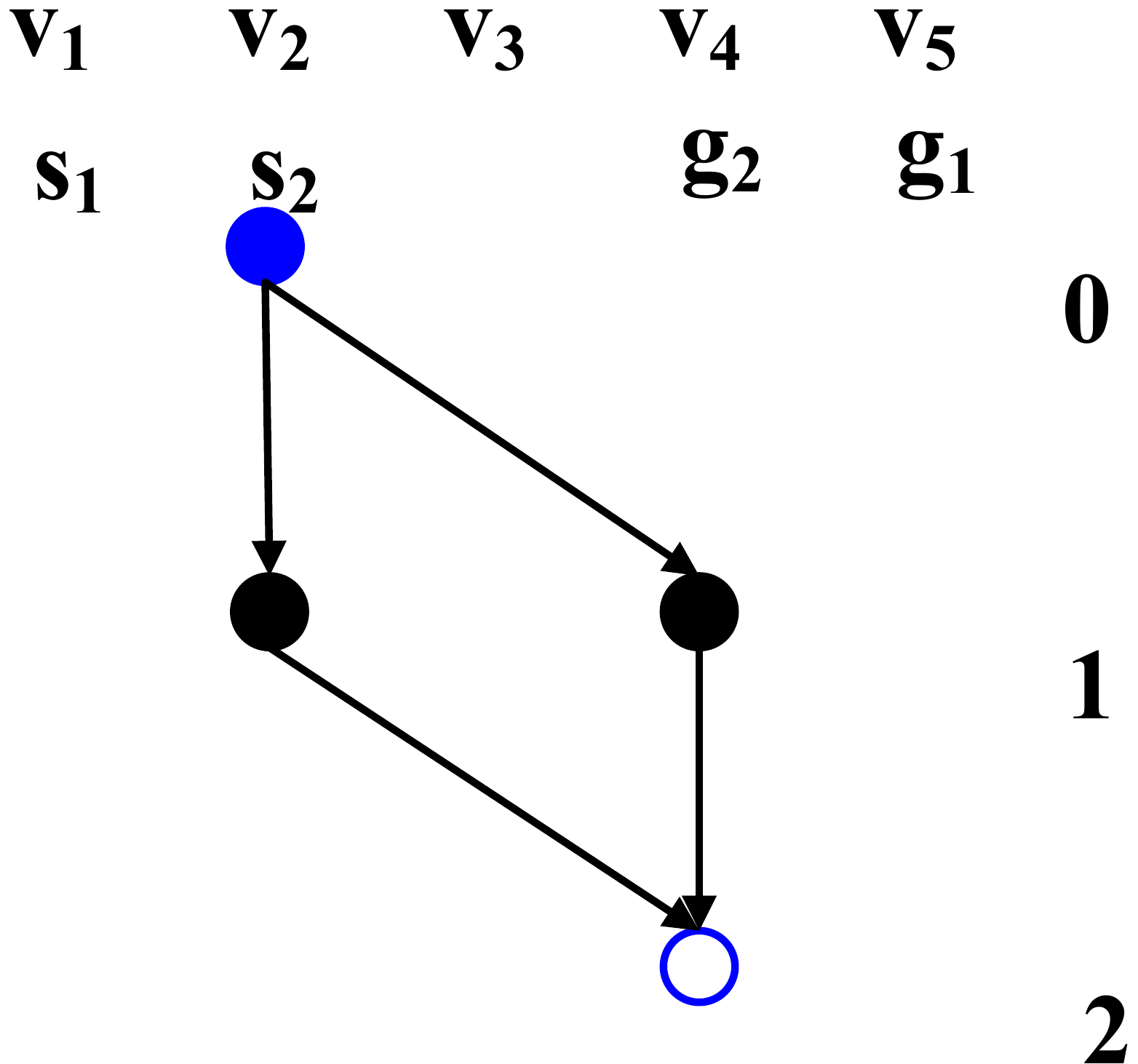}
  \caption{Optimization for the running example. Left: Abstracted flow network. Right: Reduced abstracted flow network.}\label{fig:optimization}
\end{figure}

\noindent \textbf{Abstracted Flow Network and Compact ILP Formulation} We
construct a compact ILP formulation based on an abstraction of the flow
network $\mathcal{N} = (\mathcal{V}, \mathcal{E})$ and additional
linear constraints to prevent vertex and edge collisions. We obtain the
abstracted flow network $\mathcal{N'} = (\mathcal{V'}, \mathcal{E'})$ by (1)
contracting each $(v_t^{in},v_{t}^{out})\in \mathcal{E}$ and replacing
$v_t^{in}$ and $v_{t}^{out}$ with a single vertex $v_t$ for all $v\in V$ and
$t=1\ldots \tcrit$ (and $v_{0}^{out}$ with $v_{0}$); and (2) replacing the
gadget for each $(u,v)\in E$ and each $t=0\ldots \tcrit - 1$ with a pair of
edges $(u_t, v_{t+1})$ and $(v_t, u_{t+1})$. Figure \ref{fig:optimization}
(left) shows an example. Then, we use the standard ILP formulation of this
abstracted network augmented with the constraints shown in red:
\vspace*{2mm}
{\scriptsize
\begin{flalign*}
& \text{maximize } M_{succ} = \sum_{i = 1 \ldots M} \sum_{e \in \delta^{-}((s_i)_0)} x_i[e] \text{, subject to} \nonumber\\[-1ex]
& 0 \leq \sum_{i=1}^M x_i[e] \leq 1 \quad e \in \mathcal{E'} \quad \text{(subsumed by the top red constraints)} \nonumber\\[-1ex]
& \sum_{e \in \delta^{+}(v)} x_i[e] - \sum_{e \in \delta^{-}(v)} x_i[e] = 0 \quad i=1\ldots M, v \in \mathcal{V'} \setminus \{(s_i)_0,(g_i)_{\tcrit}\} \nonumber\\[-1ex]
& \sum_{e \in \delta^{-}((s_i)_0)} x_i[e] = \sum_{e \in \delta^{+}((g_i)_{\tcrit})} x_i[e] \quad i = 1\ldots M \nonumber \\
& \color{red} \sum_{i=1\ldots M} \sum_{e \in \delta^{-}(v)}x_i[e] \leq 1 \quad v \in \mathcal{V'} \\
& \color{red} \sum_{i=1\ldots M}x_i[(u_t,v_{t+1})] + \sum_{i=1\ldots M}x_i[(v_t,u_{t+1})] \leq 1 \quad (u_t,v_{t+1}),~ (v_t, u_{t+1})\in \mathcal{E'},
\end{flalign*}}
where the 0/1 variable $x_i[e]$ represents the amount of flow of commodity
type $i$ on edge $e\in\mathcal{E'}$ and the sets $\delta^{-}(v)$ and
$\delta^{+}(v)$ contain all outgoing and incoming, respectively, edges of
vertex $v$. The top red constraints prevent vertex collisions of the form $(*,
*, v, t)$, and the bottom red constraints prevent edge collisions of the forms
$(*, *, u, v, t)$ and $(*, *, v, u, t)$.

\noindent \textbf{Reduced Abstracted Flow Network} We can remove all vertices
and edges from the abstracted flow network that are not on some path from at
least one start vertex to the corresponding goal vertex in the abstracted flow
network. This can be done by performing one complete forward breadth-first
search from each start vertex and one complete backward breadth-first search
from each goal vertex and then keeping only those vertices and edges that are
part of the search trees associated with at least one start vertex and the
corresponding goal vertex. Figure \ref{fig:optimization} (right) shows an
example. We then use the compact ILP formulation of the resulting reduced
abstracted flow network.

\noindent \textbf{Experimental Evaluation} We tested our optimal MAPF-DL
algorithm on a 2.50 GHz Intel Core i5-2450M laptop with 6 GB RAM, using CPLEX
V12.7.1 \cite{IBM2011} as the ILP solver. We randomly generated MAPF-DL
instances with different numbers of agents (ranging from 10 to 100 in
increments of 10) on $40\times40$ 4-neighbor 2D grids with deadline $\tcrit =$
50. We blocked all grid cells independently at random with 20\% probability
each. We generated 50 MAPF-DL instances for each number of agents. We placed
the start and goal vertices of each agent randomly at distance 48, 49, or
50. The following table shows the percentage of instances that could be solved
within a runtime limit of 60 seconds per instance.

{\centering
\resizebox{\columnwidth}{!}{
\begin{tabular}{|c|c|c|c|c|c|c|c|c|c|c|}
\hline
agents  & 10   & 20   & 30   & 40   & 50   & 60   & 70   & 80   & 90   & 100  \\
\hline
success rate & 100\% & 100\% & 100\% & 100\% & 98\% & 88\% & 50\% & 12\% & 0\% & 0\%\\
\hline
\end{tabular}
}}

\bibliographystyle{ACM-Reference-Format}  
\bibliography{references}  


\begin{thebibliography}{00}


\ifx \showCODEN    \undefined \def \showCODEN     #1{\unskip}     \fi
\ifx \showDOI      \undefined \def \showDOI       #1{{\tt DOI:}\penalty0{#1}\ }
  \fi
\ifx \showISBNx    \undefined \def \showISBNx     #1{\unskip}     \fi
\ifx \showISBNxiii \undefined \def \showISBNxiii  #1{\unskip}     \fi
\ifx \showISSN     \undefined \def \showISSN      #1{\unskip}     \fi
\ifx \showLCCN     \undefined \def \showLCCN      #1{\unskip}     \fi
\ifx \shownote     \undefined \def \shownote      #1{#1}          \fi
\ifx \showarticletitle \undefined \def \showarticletitle #1{#1}   \fi
\ifx \showURL      \undefined \def \showURL       #1{#1}          \fi
\providecommand\bibfield[2]{#2}
\providecommand\bibinfo[2]{#2}
\providecommand\natexlab[1]{#1}
\providecommand\showeprint[2][]{arXiv:#2}

\bibitem[\protect\citeauthoryear{Barer, Sharon, Stern, and Felner}{Barer
  et~al\mbox{.}}{2014}]%
        {ECBS}
\bibfield{author}{\bibinfo{person}{M. Barer}, \bibinfo{person}{G. Sharon},
  \bibinfo{person}{R. Stern}, {and} \bibinfo{person}{A. Felner}.}
  \bibinfo{year}{2014}\natexlab{}.
\newblock \showarticletitle{Suboptimal Variants of the Conflict-Based Search
  Algorithm for the Multi-Agent Pathfinding Problem}. In
  \bibinfo{booktitle}{{\em SoCS}}. \bibinfo{pages}{19--27}.
\newblock


\bibitem[\protect\citeauthoryear{Boyarski, Felner, Stern, Sharon, Tolpin,
  Betzalel, and Shimony}{Boyarski et~al\mbox{.}}{2015}]%
        {ICBS}
\bibfield{author}{\bibinfo{person}{E. Boyarski}, \bibinfo{person}{A. Felner},
  \bibinfo{person}{R. Stern}, \bibinfo{person}{G. Sharon}, \bibinfo{person}{D.
  Tolpin}, \bibinfo{person}{O. Betzalel}, {and} \bibinfo{person}{S.~E.
  Shimony}.} \bibinfo{year}{2015}\natexlab{}.
\newblock \showarticletitle{{ICBS}: Improved Conflict-Based Search Algorithm
  for Multi-Agent Pathfinding}. In \bibinfo{booktitle}{{\em IJCAI}}.
  \bibinfo{pages}{740--746}.
\newblock


\bibitem[\protect\citeauthoryear{Cohen, Uras, Kumar, Xu, Ayanian, and
  Koenig}{Cohen et~al\mbox{.}}{2016}]%
        {CohenUK16}
\bibfield{author}{\bibinfo{person}{L. Cohen}, \bibinfo{person}{T. Uras},
  \bibinfo{person}{T.~K.~S. Kumar}, \bibinfo{person}{H. Xu},
  \bibinfo{person}{N. Ayanian}, {and} \bibinfo{person}{S. Koenig}.}
  \bibinfo{year}{2016}\natexlab{}.
\newblock \showarticletitle{Improved Solvers for Bounded-Suboptimal Multi-Agent
  Path Finding}. In \bibinfo{booktitle}{{\em IJCAI}}.
  \bibinfo{pages}{3067--3074}.
\newblock


\bibitem[\protect\citeauthoryear{de~Wilde, ter Mors, and Witteveen}{de~Wilde
  et~al\mbox{.}}{2013}]%
        {PushAndRotate}
\bibfield{author}{\bibinfo{person}{B. de Wilde}, \bibinfo{person}{A.~W. ter
  Mors}, {and} \bibinfo{person}{C. Witteveen}.}
  \bibinfo{year}{2013}\natexlab{}.
\newblock \showarticletitle{Push and Rotate: Cooperative Multi-Agent Path
  Planning}. In \bibinfo{booktitle}{{\em AAMAS}}. \bibinfo{pages}{87--94}.
\newblock


\bibitem[\protect\citeauthoryear{Erdem, Kisa, Oztok, and Schueller}{Erdem
  et~al\mbox{.}}{2013}]%
        {erdem2013general}
\bibfield{author}{\bibinfo{person}{E. Erdem}, \bibinfo{person}{D.~G. Kisa},
  \bibinfo{person}{U. Oztok}, {and} \bibinfo{person}{P. Schueller}.}
  \bibinfo{year}{2013}\natexlab{}.
\newblock \showarticletitle{A General Formal Framework for Pathfinding Problems
  with Multiple Agents}. In \bibinfo{booktitle}{{\em AAAI}}.
  \bibinfo{pages}{290--296}.
\newblock


\bibitem[\protect\citeauthoryear{Felner, Li, Boyarski, Ma, Cohen, Kumar, and
  Koenig}{Felner et~al\mbox{.}}{2018}]%
        {FelnerICAPS18}
\bibfield{author}{\bibinfo{person}{A. Felner}, \bibinfo{person}{J. Li},
  \bibinfo{person}{E. Boyarski}, \bibinfo{person}{H. Ma}, \bibinfo{person}{L.
  Cohen}, \bibinfo{person}{T.~K.~S. Kumar}, {and} \bibinfo{person}{S. Koenig}.}
  \bibinfo{year}{2018}\natexlab{}.
\newblock \showarticletitle{Adding Heuristics to Conflict-Based Search for
  Multi-Agent Path Finding}. \bibinfo{journal}{{\em ICAPS\/}}
  (\bibinfo{year}{2018}).
\newblock


\bibitem[\protect\citeauthoryear{Felner, Stern, Shimony, Boyarski, Goldenberg,
  Sharon, Sturtevant, Wagner, and Surynek}{Felner et~al\mbox{.}}{2017}]%
        {SoCS2017Surv}
\bibfield{author}{\bibinfo{person}{A. Felner}, \bibinfo{person}{R. Stern},
  \bibinfo{person}{S.~E. Shimony}, \bibinfo{person}{E. Boyarski},
  \bibinfo{person}{M. Goldenberg}, \bibinfo{person}{G. Sharon},
  \bibinfo{person}{N.~R. Sturtevant}, \bibinfo{person}{G. Wagner}, {and}
  \bibinfo{person}{P. Surynek}.} \bibinfo{year}{2017}\natexlab{}.
\newblock \showarticletitle{Search-Based Optimal Solvers for the Multi-Agent
  Pathfinding Problem: Summary and Challenges}. In \bibinfo{booktitle}{{\em
  SoCS}}. \bibinfo{pages}{29--37}.
\newblock


\bibitem[\protect\citeauthoryear{Goldenberg, Felner, Stern, Sharon, Sturtevant,
  Holte, and Schaeffer}{Goldenberg et~al\mbox{.}}{2014}]%
        {EPEJAIR}
\bibfield{author}{\bibinfo{person}{M. Goldenberg}, \bibinfo{person}{A. Felner},
  \bibinfo{person}{R. Stern}, \bibinfo{person}{G. Sharon},
  \bibinfo{person}{N.~R. Sturtevant}, \bibinfo{person}{R.~C. Holte}, {and}
  \bibinfo{person}{J. Schaeffer}.} \bibinfo{year}{2014}\natexlab{}.
\newblock \showarticletitle{Enhanced Partial Expansion {A*}}.
\newblock \bibinfo{journal}{{\em Journal of Artificial Intelligence
  Research\/}}  \bibinfo{volume}{50} (\bibinfo{year}{2014}),
  \bibinfo{pages}{141--187}.
\newblock


\bibitem[\protect\citeauthoryear{H\"onig, Kumar, Cohen, Ma, Xu, Ayanian, and
  Koenig}{H\"onig et~al\mbox{.}}{2016a}]%
        {HoenigICAPS16}
\bibfield{author}{\bibinfo{person}{W. H\"onig}, \bibinfo{person}{T.~K.~S.
  Kumar}, \bibinfo{person}{L. Cohen}, \bibinfo{person}{H. Ma},
  \bibinfo{person}{H. Xu}, \bibinfo{person}{N. Ayanian}, {and}
  \bibinfo{person}{S. Koenig}.} \bibinfo{year}{2016}\natexlab{a}.
\newblock \showarticletitle{Multi-Agent Path Finding with Kinematic
  Constraints}. In \bibinfo{booktitle}{{\em ICAPS}}. \bibinfo{pages}{477--485}.
\newblock


\bibitem[\protect\citeauthoryear{H\"onig, Kumar, Ma, Ayanian, and
  Koenig}{H\"onig et~al\mbox{.}}{2016b}]%
        {HoenigIROS16}
\bibfield{author}{\bibinfo{person}{W. H\"onig}, \bibinfo{person}{T.~K.~S.
  Kumar}, \bibinfo{person}{H. Ma}, \bibinfo{person}{N. Ayanian}, {and}
  \bibinfo{person}{S. Koenig}.} \bibinfo{year}{2016}\natexlab{b}.
\newblock \showarticletitle{Formation Change for Robot Groups in Occluded
  Environments}. In \bibinfo{booktitle}{{\em IROS}}.
  \bibinfo{pages}{4836--4842}.
\newblock


\bibitem[\protect\citeauthoryear{IBM}{IBM}{2011}]%
        {IBM2011}
\bibfield{author}{\bibinfo{person}{IBM}.} \bibinfo{year}{2011}\natexlab{}.
\newblock \bibinfo{booktitle}{{\em {IBM ILOG CPLEX Optimization Studio CPLEX
  User's Manual}}}.
\newblock


\bibitem[\protect\citeauthoryear{Luna and Bekris}{Luna and Bekris}{2011}]%
        {PushAndSwap}
\bibfield{author}{\bibinfo{person}{R. Luna} {and} \bibinfo{person}{K.~E.
  Bekris}.} \bibinfo{year}{2011}\natexlab{}.
\newblock \showarticletitle{{Push and Swap}: Fast Cooperative Path-Finding with
  Completeness Guarantees}. In \bibinfo{booktitle}{{\em IJCAI}}.
  \bibinfo{pages}{294--300}.
\newblock


\bibitem[\protect\citeauthoryear{Ma, H\"{o}nig, Cohen, Uras, Xu, Kumar,
  Ayanian, and Koenig}{Ma et~al\mbox{.}}{2017}]%
        {MaIEEE17}
\bibfield{author}{\bibinfo{person}{H. Ma}, \bibinfo{person}{W. H\"{o}nig},
  \bibinfo{person}{L. Cohen}, \bibinfo{person}{T. Uras}, \bibinfo{person}{H.
  Xu}, \bibinfo{person}{T.~K.~S. Kumar}, \bibinfo{person}{N. Ayanian}, {and}
  \bibinfo{person}{S. Koenig}.} \bibinfo{year}{2017}\natexlab{}.
\newblock \showarticletitle{Overview: A Hierarchical Framework for Plan
  Generation and Execution in Multi-Robot Systems}.
\newblock \bibinfo{journal}{{\em IEEE Intelligent Systems\/}}
  \bibinfo{volume}{32}, \bibinfo{number}{6} (\bibinfo{year}{2017}),
  \bibinfo{pages}{6--12}.
\newblock


\bibitem[\protect\citeauthoryear{Ma and Koenig}{Ma and Koenig}{2016}]%
        {MaAAMAS16}
\bibfield{author}{\bibinfo{person}{H. Ma} {and} \bibinfo{person}{S. Koenig}.}
  \bibinfo{year}{2016}\natexlab{}.
\newblock \showarticletitle{Optimal Target Assignment and Path Finding for
  Teams of Agents}. In \bibinfo{booktitle}{{\em AAMAS}}.
  \bibinfo{pages}{1144--1152}.
\newblock


\bibitem[\protect\citeauthoryear{Ma, Koenig, Ayanian, Cohen, H\"onig, Kumar,
  Uras, Xu, Tovey, and Sharon}{Ma et~al\mbox{.}}{2016}]%
        {MaWOMPF16}
\bibfield{author}{\bibinfo{person}{H. Ma}, \bibinfo{person}{S. Koenig},
  \bibinfo{person}{N. Ayanian}, \bibinfo{person}{L. Cohen}, \bibinfo{person}{W.
  H\"onig}, \bibinfo{person}{T.~K.~S. Kumar}, \bibinfo{person}{T. Uras},
  \bibinfo{person}{H. Xu}, \bibinfo{person}{C. Tovey}, {and}
  \bibinfo{person}{G. Sharon}.} \bibinfo{year}{2016}\natexlab{}.
\newblock \showarticletitle{Overview: Generalizations of Multi-Agent Path
  Finding to Real-World Scenarios}. In \bibinfo{booktitle}{{\em IJCAI-16
  Workshop on Multi-Agent Path Finding}}.
\newblock


\bibitem[\protect\citeauthoryear{Ma, Kumar, and Koenig}{Ma
  et~al\mbox{.}}{2017a}]%
        {MaAAAI17}
\bibfield{author}{\bibinfo{person}{H. Ma}, \bibinfo{person}{T.~K.~S. Kumar},
  {and} \bibinfo{person}{S. Koenig}.} \bibinfo{year}{2017}\natexlab{a}.
\newblock \showarticletitle{Multi-Agent Path Finding with Delay Probabilities}.
  In \bibinfo{booktitle}{{\em AAAI}}. \bibinfo{pages}{3605--3612}.
\newblock


\bibitem[\protect\citeauthoryear{Ma, Li, Kumar, and Koenig}{Ma
  et~al\mbox{.}}{2017b}]%
        {MaAAMAS17}
\bibfield{author}{\bibinfo{person}{H. Ma}, \bibinfo{person}{J. Li},
  \bibinfo{person}{T.~K.~S. Kumar}, {and} \bibinfo{person}{S. Koenig}.}
  \bibinfo{year}{2017}\natexlab{b}.
\newblock \showarticletitle{Lifelong Multi-Agent Path Finding for Online Pickup
  and Delivery Tasks}. In \bibinfo{booktitle}{{\em AAMAS}}.
  \bibinfo{pages}{837--845}.
\newblock


\bibitem[\protect\citeauthoryear{Ma, Tovey, Sharon, Kumar, and Koenig}{Ma
  et~al\mbox{.}}{2016}]%
        {MaAAAI16}
\bibfield{author}{\bibinfo{person}{H. Ma}, \bibinfo{person}{C. Tovey},
  \bibinfo{person}{G. Sharon}, \bibinfo{person}{T.~K.~S. Kumar}, {and}
  \bibinfo{person}{S. Koenig}.} \bibinfo{year}{2016}\natexlab{}.
\newblock \showarticletitle{Multi-Agent Path Finding with Payload Transfers and
  the Package-Exchange Robot-Routing Problem}. In \bibinfo{booktitle}{{\em
  AAAI}}. \bibinfo{pages}{3166--3173}.
\newblock


\bibitem[\protect\citeauthoryear{Ma, Yang, Cohen, Kumar, and Koenig}{Ma
  et~al\mbox{.}}{2017}]%
        {MaAIIDE17}
\bibfield{author}{\bibinfo{person}{H. Ma}, \bibinfo{person}{J. Yang},
  \bibinfo{person}{L. Cohen}, \bibinfo{person}{T.~K.~S. Kumar}, {and}
  \bibinfo{person}{S. Koenig}.} \bibinfo{year}{2017}\natexlab{}.
\newblock \showarticletitle{Feasibility Study: Moving Non-Homogeneous Teams in
  Congested Video Game Environments}. In \bibinfo{booktitle}{{\em AIIDE}}.
  \bibinfo{pages}{270--272}.
\newblock


\bibitem[\protect\citeauthoryear{Morris, Pasareanu, Luckow, Malik, Ma, Kumar,
  and Koenig}{Morris et~al\mbox{.}}{2016}]%
        {airporttug16}
\bibfield{author}{\bibinfo{person}{R. Morris}, \bibinfo{person}{C. Pasareanu},
  \bibinfo{person}{K. Luckow}, \bibinfo{person}{W. Malik}, \bibinfo{person}{H.
  Ma}, \bibinfo{person}{S. Kumar}, {and} \bibinfo{person}{S. Koenig}.}
  \bibinfo{year}{2016}\natexlab{}.
\newblock \showarticletitle{Planning, Scheduling and Monitoring for Airport
  Surface Operations}. In \bibinfo{booktitle}{{\em AAAI-16 Workshop on Planning
  for Hybrid Systems}}.
\newblock


\bibitem[\protect\citeauthoryear{Nguyen, Obermeier, Son, Schaub, and
  Yeoh}{Nguyen et~al\mbox{.}}{2017}]%
        {GTAPF}
\bibfield{author}{\bibinfo{person}{V. Nguyen}, \bibinfo{person}{P. Obermeier},
  \bibinfo{person}{T.~C. Son}, \bibinfo{person}{T. Schaub}, {and}
  \bibinfo{person}{W. Yeoh}.} \bibinfo{year}{2017}\natexlab{}.
\newblock \showarticletitle{Generalized Target Assignment and Path Finding
  Using Answer Set Programming}. In \bibinfo{booktitle}{{\em IJCAI}}.
  \bibinfo{pages}{1216--1223}.
\newblock


\bibitem[\protect\citeauthoryear{Sharon, Stern, Felner, and Sturtevant}{Sharon
  et~al\mbox{.}}{2015}]%
        {DBLP:journals/ai/SharonSFS15}
\bibfield{author}{\bibinfo{person}{G. Sharon}, \bibinfo{person}{R. Stern},
  \bibinfo{person}{A. Felner}, {and} \bibinfo{person}{N.~R. Sturtevant}.}
  \bibinfo{year}{2015}\natexlab{}.
\newblock \showarticletitle{Conflict-Based Search for Optimal Multi-Agent
  Pathfinding}.
\newblock \bibinfo{journal}{{\em Artificial Intelligence\/}}
  \bibinfo{volume}{219} (\bibinfo{year}{2015}), \bibinfo{pages}{40--66}.
\newblock


\bibitem[\protect\citeauthoryear{Sharon, Stern, Goldenberg, and Felner}{Sharon
  et~al\mbox{.}}{2013}]%
        {DBLP:journals/ai/SharonSGF13}
\bibfield{author}{\bibinfo{person}{G. Sharon}, \bibinfo{person}{R. Stern},
  \bibinfo{person}{M. Goldenberg}, {and} \bibinfo{person}{A. Felner}.}
  \bibinfo{year}{2013}\natexlab{}.
\newblock \showarticletitle{The Increasing Cost Tree Search for Optimal
  Multi-Agent Pathfinding}.
\newblock \bibinfo{journal}{{\em Artificial Intelligence\/}}
  \bibinfo{volume}{195} (\bibinfo{year}{2013}), \bibinfo{pages}{470--495}.
\newblock


\bibitem[\protect\citeauthoryear{Silver}{Silver}{2005}]%
        {WHCA}
\bibfield{author}{\bibinfo{person}{D. Silver}.}
  \bibinfo{year}{2005}\natexlab{}.
\newblock \showarticletitle{Cooperative Pathfinding}. In
  \bibinfo{booktitle}{{\em AIIDE}}. \bibinfo{pages}{117--122}.
\newblock


\bibitem[\protect\citeauthoryear{Standley}{Standley}{2010}]%
        {ODA}
\bibfield{author}{\bibinfo{person}{T.~S. Standley}.}
  \bibinfo{year}{2010}\natexlab{}.
\newblock \showarticletitle{Finding Optimal Solutions to Cooperative
  Pathfinding Problems}. In \bibinfo{booktitle}{{\em AAAI}}.
  \bibinfo{pages}{173--178}.
\newblock


\bibitem[\protect\citeauthoryear{Standley and Korf}{Standley and Korf}{2011}]%
        {ODA11}
\bibfield{author}{\bibinfo{person}{T.~S. Standley} {and} \bibinfo{person}{R.~E.
  Korf}.} \bibinfo{year}{2011}\natexlab{}.
\newblock \showarticletitle{Complete Algorithms for Cooperative Pathfinding
  Problems}. In \bibinfo{booktitle}{{\em IJCAI}}. \bibinfo{pages}{668--673}.
\newblock


\bibitem[\protect\citeauthoryear{Sturtevant and Buro}{Sturtevant and
  Buro}{2006}]%
        {WHCA06}
\bibfield{author}{\bibinfo{person}{N.~R. Sturtevant} {and} \bibinfo{person}{M.
  Buro}.} \bibinfo{year}{2006}\natexlab{}.
\newblock \showarticletitle{Improving Collaborative Pathfinding Using Map
  Abstraction}. In \bibinfo{booktitle}{{\em AIIDE}}. \bibinfo{pages}{80--85}.
\newblock


\bibitem[\protect\citeauthoryear{Surynek}{Surynek}{2015}]%
        {Surynek15}
\bibfield{author}{\bibinfo{person}{P. Surynek}.}
  \bibinfo{year}{2015}\natexlab{}.
\newblock \showarticletitle{Reduced Time-Expansion Graphs and Goal
  Decomposition for Solving Cooperative Path Finding Sub-Optimally}. In
  \bibinfo{booktitle}{{\em IJCAI}}. \bibinfo{pages}{1916--1922}.
\newblock


\bibitem[\protect\citeauthoryear{Tovey}{Tovey}{1984}]%
        {cat1984}
\bibfield{author}{\bibinfo{person}{C.~A. Tovey}.}
  \bibinfo{year}{1984}\natexlab{}.
\newblock \showarticletitle{A Simplified {NP}-Complete Satisfiability Problem}.
\newblock \bibinfo{journal}{{\em Discrete Applied Mathematics\/}}
  \bibinfo{volume}{8} (\bibinfo{year}{1984}), \bibinfo{pages}{85--90}.
\newblock


\bibitem[\protect\citeauthoryear{Veloso, Biswas, Coltin, and Rosenthal}{Veloso
  et~al\mbox{.}}{2015}]%
        {DBLP:conf/ijcai/VelosoBCR15}
\bibfield{author}{\bibinfo{person}{M. Veloso}, \bibinfo{person}{J. Biswas},
  \bibinfo{person}{B. Coltin}, {and} \bibinfo{person}{S. Rosenthal}.}
  \bibinfo{year}{2015}\natexlab{}.
\newblock \showarticletitle{{CoBots}: Robust Symbiotic Autonomous Mobile
  Service Robots}. In \bibinfo{booktitle}{{\em IJCAI}}.
  \bibinfo{pages}{4423--4429}.
\newblock


\bibitem[\protect\citeauthoryear{Wagner and Choset}{Wagner and Choset}{2011}]%
        {DBLP:conf/iros/WagnerC11}
\bibfield{author}{\bibinfo{person}{G. Wagner} {and} \bibinfo{person}{H.
  Choset}.} \bibinfo{year}{2011}\natexlab{}.
\newblock \showarticletitle{M*: A Complete Multirobot Path Planning Algorithm
  with Performance Bounds}. In \bibinfo{booktitle}{{\em IROS}}.
  \bibinfo{pages}{3260--3267}.
\newblock


\bibitem[\protect\citeauthoryear{Wagner and Choset}{Wagner and Choset}{2015}]%
        {MStar}
\bibfield{author}{\bibinfo{person}{G. Wagner} {and} \bibinfo{person}{H.
  Choset}.} \bibinfo{year}{2015}\natexlab{}.
\newblock \showarticletitle{Subdimensional Expansion for Multirobot Path
  Planning}.
\newblock \bibinfo{journal}{{\em Artificial Intelligence\/}}
  \bibinfo{volume}{219} (\bibinfo{year}{2015}), \bibinfo{pages}{1--24}.
\newblock


\bibitem[\protect\citeauthoryear{Wang and Botea}{Wang and Botea}{2011}]%
        {WangB11}
\bibfield{author}{\bibinfo{person}{K. Wang} {and} \bibinfo{person}{A. Botea}.}
  \bibinfo{year}{2011}\natexlab{}.
\newblock \showarticletitle{{MAPP}: A Scalable Multi-Agent Path Planning
  Algorithm with Tractability and Completeness Guarantees}.
\newblock \bibinfo{journal}{{\em Journal of Artificial Intelligence
  Research\/}}  \bibinfo{volume}{42} (\bibinfo{year}{2011}),
  \bibinfo{pages}{55--90}.
\newblock


\bibitem[\protect\citeauthoryear{Wurman, D'Andrea, and Mountz}{Wurman
  et~al\mbox{.}}{2008}]%
        {kiva}
\bibfield{author}{\bibinfo{person}{P.~R. Wurman}, \bibinfo{person}{R.
  D'Andrea}, {and} \bibinfo{person}{M. Mountz}.}
  \bibinfo{year}{2008}\natexlab{}.
\newblock \showarticletitle{Coordinating Hundreds of Cooperative, Autonomous
  Vehicles in Warehouses}.
\newblock \bibinfo{journal}{{\em {AI} Magazine\/}} \bibinfo{volume}{29},
  \bibinfo{number}{1} (\bibinfo{year}{2008}), \bibinfo{pages}{9--20}.
\newblock


\bibitem[\protect\citeauthoryear{Yu and LaValle}{Yu and LaValle}{2013a}]%
        {YuLav13ICRA}
\bibfield{author}{\bibinfo{person}{J. Yu} {and} \bibinfo{person}{S.~M.
  LaValle}.} \bibinfo{year}{2013}\natexlab{a}.
\newblock \showarticletitle{Planning Optimal Paths for Multiple Robots on
  Graphs}. In \bibinfo{booktitle}{{\em ICRA}}. \bibinfo{pages}{3612--3617}.
\newblock


\bibitem[\protect\citeauthoryear{Yu and LaValle}{Yu and LaValle}{2013b}]%
        {YuLav13AAAI}
\bibfield{author}{\bibinfo{person}{J. Yu} {and} \bibinfo{person}{S.~M.
  LaValle}.} \bibinfo{year}{2013}\natexlab{b}.
\newblock \showarticletitle{Structure and Intractability of Optimal Multi-Robot
  Path Planning on Graphs}. In \bibinfo{booktitle}{{\em AAAI}}.
  \bibinfo{pages}{1444--1449}.
\newblock


\end{thebibliography}

\end{document}